\documentclass[lettersize,journal]{IEEEtran}

\usepackage{textcomp}
\usepackage{stfloats}
\usepackage{verbatim}

\usepackage{tabularx,booktabs}
\usepackage{color}
\usepackage{tabularray}
\UseTblrLibrary{diagbox}

\usepackage{cite}

\usepackage{amsmath}
\usepackage{amsfonts}
\usepackage{amssymb}
\usepackage{amsthm}

\usepackage[linesnumbered,ruled,lined]{algorithm2e}

\usepackage{array}
\usepackage{multirow}
\usepackage{makecell,rotating}       
\usepackage{float}
\usepackage{tikz}
\usetikzlibrary{3d,calc}
\usepackage{tikz-3dplot}
\usepackage{enumitem}

\ifCLASSOPTIONcompsoc
\usepackage[caption=false,font=normalsize,labelfont=sf,textfont=sf]{subfig}
\else
\usepackage[caption=false,font=footnotesize]{subfig}
\fi

\usepackage{url}

\makeatletter
\let\NAT@parse\undefined
\makeatother
\usepackage{hyperref}  

\newtheorem{definition}{Definition}
\newtheorem{example}{Example}
 
\newtheorem{corollary}{Corollary}[definition] 

\begin{document}

\title{Reliability Assessment of Information Sources Based on Random Permutation Set}

\author{
	Juntao Xu,
	Tianxiang Zhan,
	Yong Deng
	
	\thanks{This work was supported in part by the National Natural Science Foundation of China under Grant 62373078. \textit{(Corresponding author: Yong Deng.)}}
	\thanks{Juntao Xu is with the Glassgow College, University of Electronic Science and Technology of China, Chengdu 610054, China (e-mail: xujuntaouestc@hotmail.com).}
	\thanks{Tianxiang Zhan is with the Institute of Fundamental and Frontier Science, University of Electronic Science and Technology of China, Chengdu 610054, China (e-mail: zhantianxianguestc@hotmail.com).}
	\thanks{Yong Deng is with the Institute of Fundamental and Frontier Science, University of Electronic Science and Technology of China, Chengdu 610054, China, and also with the School of Medicine, Vanderbilt University, Nashville, TN 37240 USA (e-mail: dengentropy@uestc.edu.cn).}
}


\markboth{Journal of \LaTeX\ Class Files,~Vol.~14, No.~8, August~2021}%
{Shell \MakeLowercase{\textit{et al.}}: A Sample Article Using IEEEtran.cls for IEEE Journals}


\maketitle

\begin{abstract}
In pattern recognition, handling uncertainty is a critical challenge that significantly affects decision-making and classification accuracy. Dempster-Shafer Theory (DST) is an effective reasoning framework for addressing uncertainty, and the Random Permutation Set (RPS) extends DST by additionally considering the internal order of elements, forming a more ordered extension of DST. However, there is a lack of a transformation method based on permutation order between RPS and DST, as well as a sequence-based probability transformation method for RPS. Moreover, the reliability of RPS sources remains an issue that requires attention. To address these challenges, this paper proposes an RPS transformation approach and a probability transformation method tailored for RPS. On this basis, a reliability computation method for RPS sources, based on the RPS probability transformation, is introduced and applied to pattern recognition. Experimental results demonstrate that the proposed approach effectively bridges the gap between DST and RPS and achieves superior recognition accuracy in classification problems.
\end{abstract}


\section{Introduction}
\IEEEPARstart{U}{ncertain} information is ubiquitous in daily life, affecting decision-making in various domains. To address this, numerous theories have been developed, such as probability theory\cite{Probability}, intuitionistic fuzzy sets \cite{Intuitionistic_fuzzy_set}, Z-numbers \cite{Z_numberyager,ZnumberZadeh}, and Dempster-Shafer Theory (DST) \cite{Dempster,Shafer}. 

Among these, DST stands out for its ability to manage uncertainty by representing and combining evidence from multiple sources \cite{2004uncertaintyress1,2004uncertaintyress2}. Unlike probability theory, DST allows for degrees of belief distributed over sets of possibilities, making it particularly suitable for situations with reliability analysis \cite{2022reliabilityanalysisress1,2022reliabilityanalysisress2}. This flexibility enables DST to integrate disparate pieces of evidence, offering a robust approach to decision-making under uncertainty \cite{2021relclassicationress,2024desisionuncertaintyress}. Evidence theory has been further developed across various fields, including complex evidence theory \cite{Xiao2023NQMF,Xiao2023Acomplexweighted} and generalized quantum theory \cite{Xiao2023QuantumXentropy}. It has also been employed to explore the information fractal dimension to assess the complexity of mass functions \cite{Qiang2022fractal,zhan2024generalized}, as well as to introduce new entropy measures, such as Deng entropy \cite{zhao2024linearity} and generalized information entropy \cite{zhan2024generalized}. However, when handling uncertainty involving ordered information, evidence theory exhibits certain limitations. To address this issue, the Random Permutation Set (RPS) was proposed \cite{Dengrandompermutationset}. By replacing combinations with permutations, RPS introduces Permutation Event Sets (PES) and Permutation Mass Functions (PMF), which correspond to the power set and mass function in evidence theory, respectively. In subsequent research, Deng defined a method for generating PMF and effectively determining the order of fusion \cite{DengJXRPSR}; Wang further extended the orthogonal sum method within RPS \cite{Wangneworthogonalsum}; Chen proposed an RPS distance calculation based on the J-divergence measure \cite{ChenRPSdistance}; meanwhile, RPS entropy was introduced to quantify the uncertainty within RPS\cite{ChenRPSentropy}.

In pattern recognition problems, if there is significant conflict among the fused information sources, DST may produce counter-intuitive results \cite{counterintuitive, zadehparabox}. To address this issue, two common approaches have been proposed. One approach is to directly modify the combination rule, such as Yager's \cite{Yagerfusion} combination rule and Smets' unnormalized combination rule \cite{Smetsfusion}. However, such methods often compromise certain desirable properties, such as commutativity and associativity. As a result, many researchers prefer to preprocess information sources based on varying reliability, as seen in Murphy's average evidence quality method \cite{Murphyfusion} and Deng's weighted average method based on evidence distance \cite{Dengfusion}. Other approaches to reliability computation from different perspectives include Xiao's belief divergence \cite{xiaobeliefdivergence}, Liu's dissimilarity measurement \cite{Liudissimilarity}, and Jiang's correlation matrix \cite{Jiangcorrelationmatrix}.

In this paper, we conduct an in-depth analysis of RPS and develop a method for calculating support based on the internal order of elements, enabling the transformation from DST to RPS. This allows DST to overcome the constraints of order and achieve more precise mass calculations. Furthermore, considering the practical significance of permutation order in pattern recognition, we propose a Ranked Probability Transformation specifically for RPS, emphasizing the varying influence of different internal sequences on decision-making. Based on this transformation, we design a method for calculating the reliability of RPS sources. Finally, extensive numerical examples and practical applications are provided to validate the effectiveness of the proposed algorithm. The main contributions of this paper are as follows: 1. A transformation from DST to RPS is achieved, enabling DST to overcome the constraints of internal element order and allowing for more precise mass calculations. 2. A Ranked Probability Transformation for RPS is proposed, offering a more refined mass allocation for elements with different sequences. 3. The concept of decision contribution is introduced to describe the impact of individual RPS on correct decision-making, leading to the development of a method for calculating the reliability of RPS sources.

The structure of this paper is as follows. In Section \hyperref[section 2]{2}, some fundamental and related concepts are explained. Section \hyperref[section 3]{3} provides a detailed explanation of the proposed method. Section \hyperref[section 4]{4} presents numerous numerical examples and Section \hyperref[section 5]{5} gives practical applications to demonstrate the characteristics and effectiveness of the proposed algorithm. Finally, Section \hyperref[section 6]{6} concludes the paper by summarizing its key ideas.

\section{Preliminaries}
\label{section 2}
To better understand the subsequent sections, we first explain some fundamental conceptsthat are essential for this paper.

\subsection{Dempster-Shafer evidence theory}
Dempster-Shafer theory (DST) is a mathematical framework designed to handle uncertainty by representing both belief and plausibility. It has been widely used in fault diagnosis \cite{faultdiagnosis2,2022faultdiagonosisress}, clustering \cite{zhang2022bsc,Liuzhe2024clustering}, decision-making \cite{decisionmaking3,decisionmaking4}, and pattern recognition \cite{zhang2024divergence,deng2024random}, especially when dealing with incomplete or ambiguous data. One of DST's key strengths is its flexibility, as it does not require prior probabilities and provides a robust method for combining evidence. This makes it particularly effective for enhancing decision-making reliability, even in conflicting information \cite{highlyconflict1,highliconflict2}.

\begin{definition}[\textit{\textbf{Frame of discernment}}]
Let \(\Theta\) be the Frame of Discernment (FOD), which consists of a set of exhaustive and mutually exclusive elements, with each element representing a possible state of the variable, indicated by \cite{Dempster,Shafer}:
\begin{equation}
\Theta = \{x_1, x_2, x_3, \cdots, x_n\}
\label{FOD}
\end{equation}

The power set of $ \Theta $, which is denoted as $2^{\Theta}$, consists of all subsets of $ \Theta $, and can be expressed as:
\begin{equation}
2^{\Theta} = \{\emptyset, \{x_1\}, \{x_2\}, \cdots, \{x_n\}, \cdots, \{x_1, x_2, \cdots, x_n\}, \Theta\}
\label{eq:powerset}
\end{equation}
\end{definition}

\begin{definition}[\textit{\textbf{Basic probability assignment}}]
The basic probability assignment (BPA), also known as the mass function, is a mapping $2^{\theta} \rightarrow [0, 1]$, and it satisfies the following conditions \cite{Dempster,Shafer}: 
\begin{equation}
\sum_{A \in 2^{\Theta}} m(A) = 1, \quad m(\emptyset) = 0
\label{eq:bpa}
\end{equation}

where A is denoted as a focal element and it satisfies $m(A) > 0$.
\end{definition}

\begin{definition}[\textit{\textbf{Pignistic probability transformation}}]
The Pignistic Probability Transformation (PPT) successfully converts the Basic Probability Assignment (BPA) into probabilities for the final decision-making stage \cite{PPT1}. Its core concept is to evenly distribute the BPA of a focal element containing multiple elements among each of its internal elements, thereby ensuring fairness and objectivity. For a given FOD $ \Theta $ the PPT is defined as follows:
\begin{equation}
\text{BetP}(x_i) = \sum_{x_i \in A} \frac{m(A)}{|A|}, \quad A \in 2^{\Theta}
\label{eq:ppt}
\end{equation}

where $|A|$ is the number of elements in focal $A$.
\end{definition}

\begin{definition}[\textbf{\textit{Discounting rules}}]
In pattern recognition, different sources of evidence have varying degrees of reliability. Therefore, the discounting rule is introduced to redistribute the BPA. Given a FOD $ \Theta $ with its corresponding mass function $ m(\cdot) $, the discounting rule is defined by \cite{Shafer}:
\begin{equation}
m'(A) = 
\begin{cases} 
m(A) \cdot \beta, & A \in 2^{\Theta} \ \text{and} \ A \neq \Theta \\ 
m(A) \cdot \beta + (1 - \beta), & A = \Theta
\end{cases}
\label{eq:discount}
\end{equation}

where $\beta$ represents evidential reliability.
\end{definition}

\subsection{Random permutation set theory}
The Random Permutation Set (RPS) is an innovative extension of DST that additionally considers the potential order of elements within a focal element. This internal order can represent varying importance, recognition of internal possibilities, and other characteristics. Based on this internal predefined order, more detailed reasoning and decision-making can be achieved. Some fundamental concepts related to RPS will be explained below.

\begin{definition}[\textbf{\textit{Permutation Event Space}}] 
Considering a FOD $ \Theta = \{x_1, x_2, x_3, \cdots, x_n\} $ with an internal order, the corresponding PES is expressed as \cite{Dengrandompermutationset}:
\begin{equation}
\begin{aligned}
\text{PES}(\Theta) = & \left\{ A_i^j \mid i = 0, 1, 2, \cdots, n; \ j = 0, 1, 2, \cdots, P(n, i) \right\} \\
= & \left\{ \emptyset, (x_1), (x_2), \cdots, (x_n), (x_1, x_2), (x_2, x_1), \cdots, \right. \\
& \left( (x_{n-1}, x_n), (x_n, x_{n-1}), \cdots, (x_1, x_2, \cdots, x_n), \right. \\
& \left. \cdots, (x_n, x_{n-1}, \cdots, x_1) \right\}
\end{aligned}
\end{equation}

where $i$ is the number of elements in one focal and $P(n, i)$ is the number of permutation in focal with $i$ elements, calculated by $P(n, i) = \frac{n!}{(n - i)!}$. $j$ represents the index of the permutation for a given number of elements. For each $ A_i^j $ in the PES, it is referred to as a Permutation Event (PE).
\end{definition}

\begin{definition}[\textbf{\textit{Random permutation set}}]
Given a Frame of Discernment (FOD) $ \Theta = \{x_1, x_2, x_3, \cdots, x_n\} $, its RPS is a series of pair consisting of elements from the PES \cite{Dengrandompermutationset}:
\begin{equation}
RPS(\Theta) = \left\{ \langle A, \mu(A) \rangle \mid A \in PES(\Theta) \right\}
\label{eq:RPS}
\end{equation}

where $ \mu(A) $ is defined as the Permutation Mass Function (PMF), which is a mapping $ PES(\Theta) \rightarrow [0, 1] $, and it satisfies:
\begin{equation}
\mu(\emptyset) = 0, \quad \sum_{A \in PES(\Theta)} \mu(A) = 1
\label{eq:PMFcondition}
\end{equation}

RPS introduces an internal order of elements within a focal element on the basis of the BPA. When a focal element of the BPA contains multiple elements, their permutation order can potentially reflect their relative importance, effectively subdividing the original focal element based on an internal ranking. For example, given the FOD $ \Theta = \{x_1, x_2, x_3\} $, the corresponding subdivision of PEs is illustrated in \hyperref[fig:BPA-RPS]{Fig. 1}.

\begin{figure}[ht]
    \centering
    \includegraphics[width=0.4\textwidth]{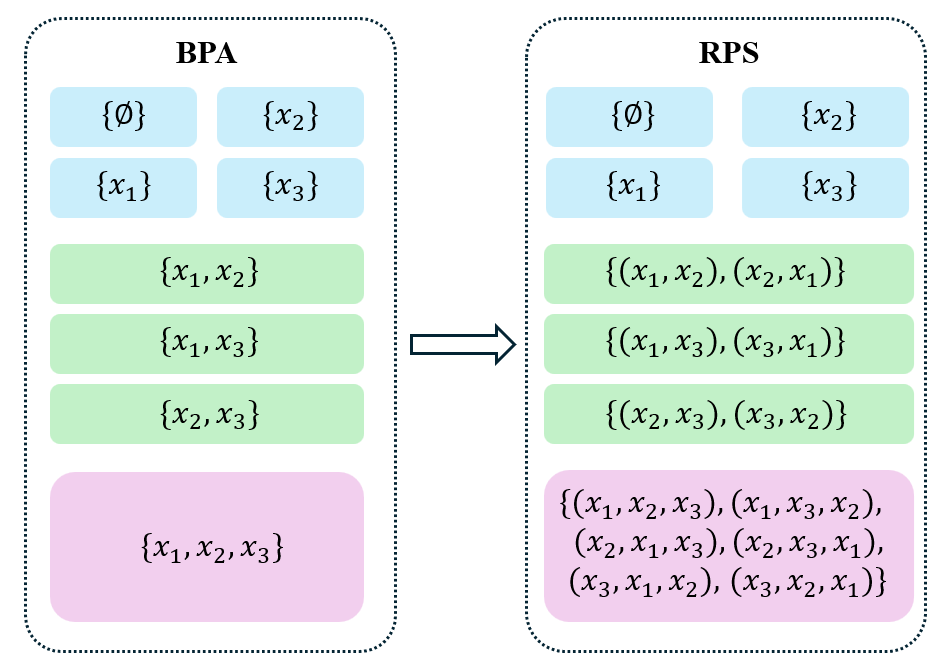}  
    \caption{An illustration of the subdivision of PEs for the given FOD}
    \label{fig:BPA-RPS}
\end{figure}

If the order is ignored, the Permutation Event Space will degenerate into the power set of $ \Theta $; the Permutation Mass Function will degenerate into the mass function.
\end{definition}

\begin{definition}[\textit{\textbf{Intersection of permutation events}}]
Let $ A $ and $ B $ be two permutation events in the Permutation Event Space (PES) of $ \Theta $. The calculation of the Left Intersection (LI) and Right Intersection (RI) of $ A $ and $ B $ are defined as follows \cite{Dengrandompermutationset}:
\begin{equation}
A \overset{\leftarrow}{\cap} B = A \setminus \{w \mid w \in A, w \notin B\} \quad (\text{LI})
\label{eq:LI}
\end{equation}

\begin{equation}
A \overset{\rightarrow}{\cap} B = B \setminus \{w \mid w \in B, w \notin A\} \quad (\text{RI})
\label{eq:RI}
\end{equation}

where $ P \setminus Q $ denotes removing $ Q $ from $ P $ while preserving the order of the remaining elements in $ P $.
\end{definition}

\begin{definition}[\textit{\textbf{Orthogonal sum of permutation mass functions}}]
Let $ \mu_1 $ and $ \mu_2 $ be two PMFs. Their Right Orthogonal Sum (ROS) is defined as \cite{Dengrandompermutationset}:
\begin{equation}
\begin{aligned}
\mu^R(A) &\equiv \mu_1 \overset{\rightarrow}{\oplus} \mu_2 (A) = \\
& \begin{cases}
\frac{1}{1 - \overset{\rightarrow}{K}} \cdot \sum_{B \overset{\rightarrow}{\cap} C = A} \mu_1(B) \cdot \mu_2(C), & A \neq \emptyset \\
0, & A = \emptyset
\end{cases}    
\end{aligned}
\label{ROS}
\end{equation}

where $A, B, C \in \text{PES}(\Theta)$, $ \overset{\rightarrow}{\cap} $ denotes the right intersection, and $ \overset{\rightarrow}{K} $ is defined as:

\begin{equation}
\overset{\rightarrow}{K} = \sum_{B \overset{\rightarrow}{\cap} C = \emptyset} \mu_1(B) \cdot \mu_2(C)
\label{eq:RK}
\end{equation}
\quad If the order is swapped, the Left Orthogonal Sum (LOS) of \( \mu_1 \) and \( \mu_2 \) is defined as:
\begin{equation}
\begin{aligned}
\mu^L(A) &\equiv \mu_1 \overset{\leftarrow}{\oplus} \mu_2 (A) = \\
&\begin{cases}
\frac{1}{1 - \overset{\leftarrow}{K}} \cdot \sum_{B \overset{\leftarrow}{\cap} C = A} \mu_1(B) \cdot \mu_2(C), & A \neq \emptyset \\
0, & A = \emptyset
\end{cases}    
\end{aligned}
\label{LOS}
\end{equation}

where $A, B, C \in \text{PES}(\Theta)$, $ \overset{\leftarrow}{\cap} $ denotes the right intersection, and $ \overset{\leftarrow}{K} $ is defined as:

\begin{equation}
\overset{\leftarrow}{K} = \sum_{B \overset{\leftarrow}{\cap} C = \emptyset} \mu_1(B) \cdot \mu_2(C)
\label{eq:LK}
\end{equation}
\end{definition}

\begin{definition}[\textbf{\textit{Ordered probability transformation}}]
To convert the PMF into a probability distribution, the Ordered Probability Transformation (OPT) is proposed. Given an $RPS(\Theta) = \left\{ \langle A, \mu(A) \rangle \mid A \in PES(\Theta) \right\}$, OPT is represented as \cite{DengJXRPSR}:
\begin{equation}
\begin{aligned}
    \text{OPT}(x_i) = \mu(\{x_i\}) + \sum_{x_i \in A \in PES(\Theta)} \frac{\mu(A)}{|A| - 1} \mid \\
    \text{Last}(A) \neq x_i, \ |A| > 1
\end{aligned}
\label{eq:OPT}
\end{equation}

where $ \text{Last}(A) \neq x_i $ indicates that if $ x_i $ is the last element of $ A $, meaning it is the least important in the internal order, it will be ignored during the probability allocation process. The key of OPT is to redistribute the mass of multi-element PEs evenly while ignoring the least significant elements.
\end{definition}

\begin{definition}[\textbf{\textit{RPS discounting rule}}]
Given an $RPS(\Theta) = \left\{ \langle A, \mu(A) \rangle \mid A \in PES(\Theta) \right\}$ with reliability $\alpha $, its discounting rule is defined as \cite{DengJXRPSR}:
\begin{equation}
\mu'(A) =
\begin{cases}
\mu(A) \cdot \alpha, & |A| = 1 \\
\mu(A) \cdot \alpha + \frac{1 - \alpha}{\text{Perm}(|\Theta|) - |\Theta| - 1}, & |A| > 1
\end{cases}
\label{eq:RPSdiscount}
\end{equation}

where $ \text{Perm}(k) = \sum_{i=0}^{k} P(k, i) $ is the total number of all permutations in the PES of that RPS. Considering the impact of internal order, the discounting of the RPS evenly distributes the uncertainty (1 - reliability) among all possible permutations to ensure fairness.
\end{definition}

\begin{definition}[\textbf{\textit{BPA generation using Gaussian discriminant Model}}]
Assume that the object to be recognized is $ O_i $ with $j$ features, and its potential labels are $ \{\theta_1, \theta_2, \dots, \theta_n\} $. During the training process, let the sample size be $ N $. For the recognized object $ O_i $, its membership degree based on the Gaussian Discriminant Model is given by \cite{DengJXRPSR}:

\begin{equation}
f^{j}\{O_i\}(\theta_n) = \frac{1}{\sqrt{2 \pi (\sigma^j)^2}} \cdot \exp\left[- \frac{(x_i^j - \bar{x}^j)^2}{2 (\sigma^j)^2}\right] \bigg| {x_i \leftarrow \theta_n}
\label{Gaussian membership}
\end{equation}

where
\begin{equation}
\bar{x}^j = \frac{1}{N} \cdot \sum_{i=1}^{N} {x_i^j} \bigg| {x_i \leftarrow \theta_n}
\label{xba}
\end{equation}

\begin{equation}
\sigma^j = \sqrt{\frac{1}{N-1} \cdot \sum_{i=1}^{N} (x_i^j - \bar{x}^j)^2 } \bigg| {x_i \leftarrow \theta_n}
\label{sigma}
\end{equation}

$ f^{j}{O_i}(\theta_n) $ represents the membership of $ O_i $ to $ \theta_n $ based on the $ j $-th attribute, and $x_i \leftarrow \theta_n$ represents the value of the $ j $-th attribute for all samples labeled as $ \theta_n $ during the training process.

After obtaining the membership values of $O_i$ for all labels, these membership values are normalized as follows:
\begin{equation}
\hat{f}^{j}\{O_i\}(\theta_n) = \frac{f^{j}\{O_i\}(\theta_n)}{\sum_{i=1}^{N} f^{j}\{O_i\}(\theta_n)}
\label{normalized membership}
\end{equation}

The generation of BPA is based on the normalized membership degrees for each label defined as:
\begin{equation}
m_j (\theta_n \cup \theta_{\text{greater}}) = \hat{f}^{j}\{O_i\}(\theta_n)
\label{bpa_generation}
\end{equation}

where $ \theta_{\text{greater}} $ represents the union of all other labels whose normalized membership is greater than $ \theta_n $.
\end{definition}

\section{Proposed methods}
\label{section 3}
In this section, the transformation of DST and RPS will be introduced by incorporating the concept of support. Additionally, considering the intrinsic order within the RPS, a new OPT algorithm will be proposed. Furthermore, the calculation of the reliability of the source in RPS will also be discussed.

\begin{definition}[\textbf{\textit{Internal orders ranking}}]
Given a FOD $ \Theta = \{x_1, x_2, x_3, \cdots, x_n\} $, for any $ A \in \text{PES}(\Theta) $, the internal order ranking of $ A $ is defined as:
\begin{equation}
\beta (A) = \left\{ \beta_1, \beta_2, \beta_3, \dots, \beta_n \mid n = |A| \right\}, \quad A \in \text{PES}(\Theta)
\label{eq:IOR}
\end{equation}

where $ \beta $ is the order of element in $A$. By utilizing internal order ranking, each element in the PEs can be represented by its corresponding $ \beta $, which facilitates the subsequent mass allocation.
\end{definition}

\begin{example}
Given a FOD $Theta = \{D, N, A\}$, its Permutation Events and corresponding internal order rankings are shown in \hyperref[table1]{Table 1}.

\begin{table}[ht]
\centering
\caption{PEs and corresponding internal order rankings.}
\begin{tabular}{llllllll}
\toprule[1.5pt] 
 & $\beta_1$ & $\beta_2$ & $\beta_3$ & & $\beta_1$ & $\beta_2$ & $\beta_3$ \\
\cmidrule(lr){2-4} \cmidrule(lr){6-8} 
(D)    & D & 0 & 0 & (D, N, A) & D & N & A \\
(N)    & N & 0 & 0 & (D, A, N) & D & A & N \\
(A)    & A & 0 & 0 & (N, D, A) & N & D & A \\
(D, N) & D & N & 0 & (N, A, D) & N & A & D \\
(N, D) & N & D & 0 & (A, D, N) & A & D & N \\
(D, A) & D & A & 0 & (A, N, D) & A & N & D \\
(A, D) & A & D & 0 & & & & \\
(N, A) & N & A & 0 & & & & \\
(A, N) & A & N & 0 & & & & \\
\bottomrule[1.5pt] 
\end{tabular}
\label{table1}
\end{table}

\end{example}

\begin{definition}[\textbf{\textit{Ordered support degree}}]
Given a FOD $ \Theta = {x_1, x_2, x_3, \cdots, x_n} $, for any $ A \in PES(\Theta) $, the order support of $ A $ is defined as:
\begin{equation}
\text{Sord}(A) = \prod_{i=1}^{|A|} \frac{\text{BetP}(\beta_i)}{\sum_{j=i}^{|A|} \text{BetP}(\beta_j)}
\label{eq:osd}
\end{equation}

where $ \text{BetP}(\beta_i) $ represents the probability distribution of element $ \beta_i $ obtained after applying the Pignistic Probability Transformation (PPT) on the initial BPA.
\end{definition}

\begin{corollary}
Since the RPS additionally considers the impact of order compared to DST, for different $ A \in PES(\Theta) $, if they contain the same number of elements, their order support satisfies:
\begin{equation}
\sum_{|A|=g} \text{Sord}(A) = 1, \quad g \in \{1, 2, 3, \dots, n\}
\label{eq:sumsord1}
\end{equation}

where $g$ denotes the number of elements contained in $A$.
\end{corollary}

\begin{corollary}
If $ A $ contains only one element, then $ \text{Sord}(A) $ is always equal to 1.
\end{corollary}

\begin{example}
Given a FOD $ \Theta = \{D, N, A\} $, assume that the probability distributions obtained after the Pignistic Probability Transformation are:
\begin{equation}
\text{BetP}(D) = 0.2, \quad \text{BetP}(N) = 0.3, \quad \text{BetP}(A) = 0.5
\end{equation}

\begin{itemize}[label=$\circ$]  
    \item For a PE containing only one element, \( \text{Sord} \) is always equal to 1.

    \item For a PE containing two elements, such as \( (N, D) \), its \( \text{Sord} \) is:
    \begin{equation}
    \text{Sord}(N, D) = \frac{\text{BetP}(\beta_1)}{\text{BetP}(\beta_1) + \text{BetP}(\beta_2)} \cdot \frac{\text{BetP}(\beta_2)}{\text{BetP}(\beta_2)} = 0.6
    \end{equation}
    \quad where $\beta_1$, $\beta_2$ denote $N$ and $D$ in $(N, D)$ respectively.
    \item For a PE containing three elements, such as \( (A, D, N) \), its \( \text{Sord} \) is:
\begin{align}
\text{Sord}(A, D, N) &= \frac{\text{BetP}(\beta_1)}{\text{BetP}(\beta_1) + \text{BetP}(\beta_2) + \text{BetP}(\beta_3)} \notag \\
& \cdot \frac{\text{BetP}(\beta_2)}{\text{BetP}(\beta_2) + \text{BetP}(\beta_3)} \notag \\
&\cdot \frac{\text{BetP}(\beta_3)}{\text{BetP}(\beta_3)} = 0.2
\end{align}

where $\beta_1$, $\beta_2$ and $\beta_3$ denote $A$, $D$ and $N$ in $(A, D, N)$ respectively.
\end{itemize}
\end{example}

\begin{definition}[\textbf{\textit{RPS transformation}}]
Given a FOD $ \Theta $, for any $ A \in PES(\Theta) $, the RPS transformation is defined as:
\begin{equation}
\mu(A) = m(\text{Element}(A)) \cdot \text{Sord}(A)
\label{RPST}
\end{equation}

where \text{Element}(A) represents $ A $ in the original DST's BPA, disregarding its order.

Noting that RPS takes into account the additional factor of order compared to DST, the PMF essentially represents a finer division of the BPA. The rule for this finer division must consider the arrangement of the elements.
\end{definition}

\begin{example}
Given a FOD $ \Theta = \{D, N, A\} $, which satisfies:

\begin{equation}
\begin{aligned}
m(D) &= 0.1, \quad m(N) = 0.2, \quad m(A) = 0.2, \\
m(N, A) &= 0.2, \quad m(D, N, A) = 0.3
\end{aligned}
\end{equation}

\begin{enumerate}[label=\textbf{Step \arabic*:}, leftmargin=1.5cm, labelwidth=1.5cm]
    \item Calculate the BetP corresponding to the objects in the original BPA based on the Pignistic probability transformation.
    \item Calculate ordered support degree for PEs according to BetP.
    \item Calculate transformed PMF based on Pord and original BPA.
\end{enumerate}

The BetP values for each object, the initial BPA, and the resulting PEs and PMF after the RPS transformation are presented in 
\begin{table}[H]
\centering
\caption{The BPA of DST and the corresponding PMF after RPS transformation.}
\resizebox{\linewidth}{!}{
\begin{tabular}{c c|c c|c c @{\hskip 15pt} || @{\hskip 15pt} c c|c c}
\toprule[1.5pt]
object & BetP($\cdot$) & focal & m($\cdot$) & PE & $\mu(\cdot)$ & focal & m($\cdot$) & PE & $\mu(\cdot)$ \\ \hline
D      & 0.2           & (D)   & 0.1        & (D)  & 0.1          & (D,N,A)   & 0.3 & (D,N,A)   & 0.03 \\
N      & 0.4           & (N)   & 0.2        & (N)  & 0.2          &           &     & (D,A,N)   & 0.03 \\
A      & 0.4           & (A)   & 0.2        & (A)  & 0.2          &           &     & (N,D,A)   & 0.04 \\
       &               & (N,A) & 0.2        & (N,A)& 0.1          &           &     & (N,A,D)   & 0.08 \\
       &               &       &            & (A,N)& 0.1          &           &     & (A,D,N)   & 0.04 \\
       &               &       &            &      &              &           &     & (A,N,D)   & 0.08 \\ \bottomrule[1.5pt]
\end{tabular}
}
\label{table2}
\end{table}

\end{example}

\begin{definition}[\textbf{\textit{Ranked probability transformation}}]
Given a FOD $ \Theta = \{x_1, x_2, x_3, \cdots, x_n\} $, the Ranked Probability Transformation (RPT) for the corresponding RPS is defined as:

\begin{equation}
\text{Rpt}(x_i) = \sum_{A \in PES(\Theta)} \frac{e^{-\frac{\lambda}{1-\lambda} \cdot \text{rank}(x_i)}}{\sum_{x_i \in A} e^{-\frac{\lambda}{1-\lambda} \cdot \text{rank}(x_i)}} \cdot m(A)
\label{RPT}
\end{equation}

where $ \text{rank}(x_i) $ denotes the position of $x_i$ among the elements of $A$.

Considering that in RPS, elements ranked higher should receive a larger portion of the PMF, a dispersion factor is introduced to achieve this effect. The factor $e^{-\frac{\lambda}{1-\lambda}}$ serves as the dispersion factor, enabling the redistribution of PMF according to different weights based on the rank of elements. The value of $\lambda$ ranges from [0,1], with larger values of $\lambda$ causing the rank to have a greater effect on PMF distribution, leading to more dispersed PMF. When $\lambda = 0$, the RPT reduces to the standard PPT. In this paper, $\lambda$ is set to a default value of 0.67.
\end{definition}

\begin{definition}[\textbf{\textit{Decision contribution}}]
In the training phase of pattern recognition, given a FOD $ \Theta = \{x_1, x_2, x_3, \cdots, x_n\} $, the decision contribution of a PRS source is defined as:
\begin{equation}
dc_k^j = Rpt_k^j(x^*) - \frac{\sum_{x_i \in \{\Theta - x^*\}} Rpt_k^j(x_i)}{|\Theta - x^*|}
\label{dc}
\end{equation}

where $x^*$ denotes the correct classification of the recognized object, $j$ represents the index of the recognized sample, $k$ is the index of the RPS source, and $ |\Theta - \{x^*\}| $ denotes the number of the remaining incorrect classes.

It is noted that $dc$ not only takes into account the Rpt of the correct classification, but also is related to the Rpt of incorrect classifications. If the average Rpt for the other incorrect categories is greater than that of the correct one, the RPS source is considered to have made a negative contribution to the decision.
\end{definition}

\begin{definition}[\textbf{\textit{RPS reliability calculation}}]
In pattern recognition, given $ k $ RPS sources $ \{ RPS_1, RPS_2, \dots, RPS_k \} $, their reliability is defined as:
\begin{equation}
\begin{aligned}
R_k = \frac{DC_k - \min \left\{ \widehat{DC} \right\}}{\max \left\{ \widehat{DC} \right\} - \min \left\{ \widehat{DC} \right\}}, \\ \widehat{DC} = \{DC_1, DC_2, \dots, DC_k\}
\end{aligned}
\label{Rel}
\end{equation}

where
\begin{equation}
DC_k = \sum_{j=1}^{N} dc_k^j
\label{sumofdc}
\end{equation}

$ N $ is the sum of sample number used for training.

Note that the calculation of reliability is based on the relative contribution of different RPS sources to correct decision-making. Based on Eq.\hyperref[Rel]{(32)}, we know that the RPS source contributing the most to correct decisions has a reliability of 1, while the source contributing the least is considered entirely unreliable, with a reliability of 0.
\end{definition}

\begin{algorithm}[h]
\caption{Weight calculation based on random permutation set transformation}
\KwIn{the FOD $\Theta=\left\{x_1, x_2, x_3 \cdots x_n\right\}$, a set of objects to be identified $O=\left\{o_1, o_2, \ldots o_N\right\}$, a set of evidence sources $S=\left\{S_1, S_2, \ldots S_k\right\}$, BPAs of objects in object set $O_N$ from one evidence source $S_k$}
\KwOut{ RPS source reliability \( R_k \)}
Initialize \( R_j \) to 0 \;
\For{each source $S_j \in\left\{S_1, S_2 \cdots S_k\right\}$}
{
    \For{each subject $o_i \in\left\{o_1, o_2 \cdots o_N\right\}$}
    {
        Take a BPA for an object $o_i$ from Source $S_j$ \;
        Perform the RPS transformation of the BPA based on Eq.(17-18, 23) to obtain the corresponding RPS \;
        Calculate the decision contribution $dc_k^j$ based on Eq.(25-26) \;
    }
    Calculate the total decision contribution \( DC_k \) based on Eq.(28) \;
}
Determine the RPS source reliability \( R_k \) according to Eq.(27)\;
\Return $R_k$
\end{algorithm}

\section{Numerical examples}
\label{section 4}
\begin{example}
Given a FOD $ \Theta = \{x_1, x_2, x_3\} $, during the training phase, assume the correct label exists in the form of $ RPS^* $. Given an $ RPS_1 $, it satisfies the following condition with $ RPS^* $:
\begin{equation}
\operatorname{RPS}_1=\left\{\left\langle(x_1),0.4\right\rangle,\quad \left\langle(x_1,x_2), 0.2\right\rangle,\quad \left\langle(A), 0.4\right\rangle\right\}
\end{equation}

\begin{equation}
\operatorname{RPS^*}=\left\{\left\langle(x_1),1\right\rangle\right\}
\end{equation}

where $ A \in PES(\Theta) $, and the order of $ A $ is disregarded, then $ RPS_1 $ will degenerate into the following mass function:
\begin{equation}
m(x_1) = 0.4, \quad m(x_1, x_2) = 0.2, \quad m(A) = 0.4
\end{equation}

To better understand the relationship between the proposed distance calculation method and the internal order, in this example, the proposed method will be compared with the widely used J distance. The specific comparison is shown in \hyperref[table_comparison]{Table 3}. For convenience, the composition of A is represented using indices on the graph, as shown in \hyperref[rel1]{Fig. 2}.

\begin{figure}[h]
    \centering
    \includegraphics[width=\linewidth]{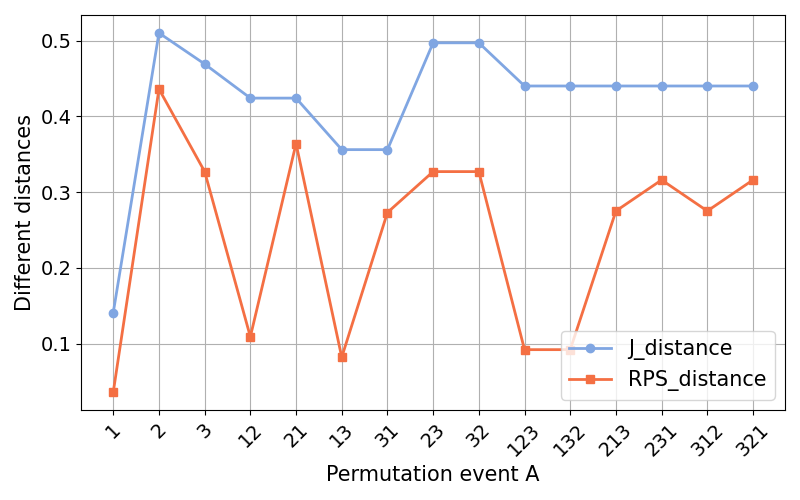} 
    \caption{Different distance under different A.}
    \label{rel}
\end{figure}

\begin{table}[H]
\centering
\caption{Comparison of J distance and RPS distance.}
\resizebox{\linewidth}{!}{
\begin{tabular}{c | c|c @{\hskip 15pt} || @{\hskip 15pt} c | c|c}
\toprule[1.5pt]
A   & J distance & RPS distance & A   & J distance & RPS distance \\ \hline
$x_1$  & 0.141 & 0.036 & ($x_2$, $x_3$) & 0.497 & 0.327 \\
$x_2$  & 0.510 & 0.436 & ($x_3$, $x_2$) & 0.497 & 0.327 \\
$x_3$  & 0.469 & 0.327 & ($x_1$, $x_2$, $x_3$) & 0.440 & 0.092 \\
($x_1$, $x_2$)  & 0.424 & 0.109 & ($x_1$, $x_3$, $x_2$) & 0.440 & 0.092 \\
($x_2$, $x_1$)  & 0.424 & 0.364 & ($x_2$, $x_1$, $x_3$) & 0.440 & 0.275 \\
($x_1$, $x_3$)  & 0.356 & 0.082 & ($x_2$, $x_3$, $x_1$) & 0.440 & 0.316 \\
($x_3$, $x_1$)  & 0.356 & 0.273 & ($x_3$, $x_1$, $x_2$) & 0.440 & 0.275 \\
                &       &       & ($x_3$, $x_2$, $x_1$) & 0.440 & 0.316 \\ \bottomrule[1.5pt]
\end{tabular}
}
\label{table_comparison}    
\end{table}

There are several points worth noting from \hyperref[rel]{{Fig. 2}} and \hyperref[table_comparison]{Table 3}:

1. When A contains multiple elements, for J distance \cite{Jdistance}, which represents distance based on DS theory, the internal order of elements is not considered, so the distance remains unchanged. However, the proposed method considers the RPS, so the distance changes with the internal order of the elements.

2. It is noticeable that whenever the first element of A is $x_1$ (the correct label), the J distance shows a larger difference compared to the RPS distance. This is because in RPS, having $x_1$ in the first position indicates that $A$ is most likely the target being recognized. According to the proposed Ranked Probability Transformation, $x_1$ will be assigned a higher proportion in the probability transformation. In DS theory, since there is no distinction in the order of elements, some information is lost, resulting in a larger distance.

3. When performing internal comparisons of RPS, it is found that the closer $x_1$ is to the front of A, the smaller the distance between $RPS_1$ and $RPS^*$. When the position of $x_1$ is the same and the elements of A are the same, the distance between RPS does not change with the positions of $x_2$ and $x_3$. This is because the reliability calculation of the proposed method is based on the difference between the Rpt that supports the correct recognition result and the Rpt that does not support it, while the internal composition of the opposing part has little impact on the reliability.
\end{example}

\begin{example}
Given an FOD $ \Theta = \{x_1, x_2, x_3\} $, $ m_1 $ and $ m^* $ satisfy: 
\begin{equation}
\begin{aligned}
 m_1(x_1) &= \eta, \quad m_1(x_3) = 0.7 - \eta, \\
 m_1(x_2,x_3) &= 0.2, \quad m_1(x_1,x_2,x_3) = 0.1   
\end{aligned}
\end{equation}

\begin{equation}
m^*(x_1) = 1
\end{equation}

where $ \eta \in [0, 0.7] $, it is used to modify the mass function $ m_1 $, with a step size of 0.01 for each change. 

Since the reliability of $m_1$ is relative, in this example, we use $m^*$ and $m_2$ (where $m_2(x_2,x_3) = 1$) to determine the upper and lower bounds of $m_1$. For comparative analysis, we employ Liu’s dissimilarity measurement \cite{Liudissimilarity}, Jiang’s correlation coefficient \cite{Jiangcorrelationmatrix}, Lefevre’s adapted conflict \cite{Lefevredistance}, and Jousselme et al.’s distance \cite{Jdistance} to compare with the proposed reliability calculation method. The specific results are shown in \hyperref[rel1]{Fig. 3}.

\begin{figure}[h]
    \centering
    \includegraphics[width=\linewidth]{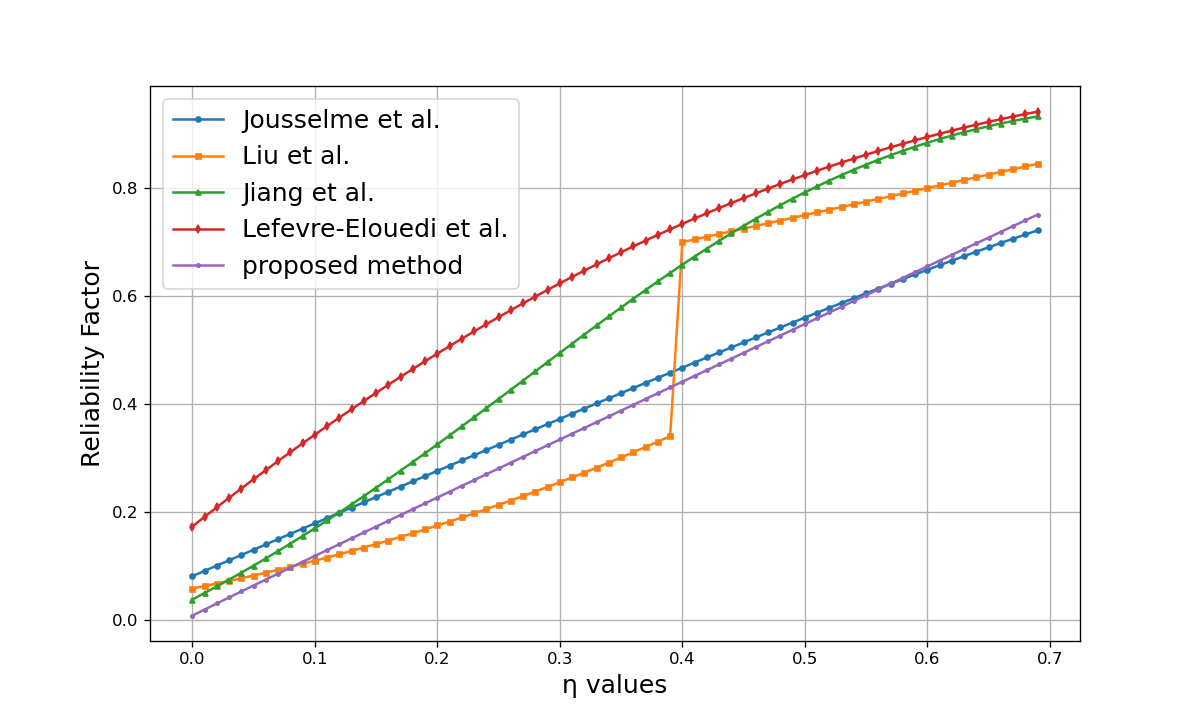} 
    \caption{The reliability factor of information source under different $\eta$.}
    \label{rel1}
\end{figure}

From \hyperref[rel1]{Fig. 3}, with the uniform increase of $ \eta $, the reliability of $ m_1 $ also steadily increases, and the evidence source, after the RPS transformation, exhibits an approximately linear change in reliability.
\end{example}

\begin{example}
Given a FOD \( \Theta = \{x_1, x_2, x_3\} \), \( m_1 \) and \( m^* \) are respectively defined as follows:
\begin{equation}
\begin{aligned}
m_1(x_1) &= 0.1, \quad m_1(x_3) = \eta, \\
m_1(x_2,x_3) &= 0.7 - \eta, \quad m_1(x_1,x_2,x_3) = 0.2
\end{aligned}
\end{equation}

\begin{equation}
m^*(x_1) = 1
\end{equation}

where $ \eta \in [0, 0.7] $, with a step size of 0.01. Similarly, since the reliability of $m_1$ is relative, in this example, we also use $m^*$ and $m_2$, where $m_2(x_2, x_3) = 1$, to determine the upper and lower limits of $m_1$. The comparison results of the proposed reliability calculation method with other DS-based methods are shown in \hyperref[rel2]{Fig. 4}.

\begin{figure}[h]
    \centering
    \includegraphics[width=\linewidth]{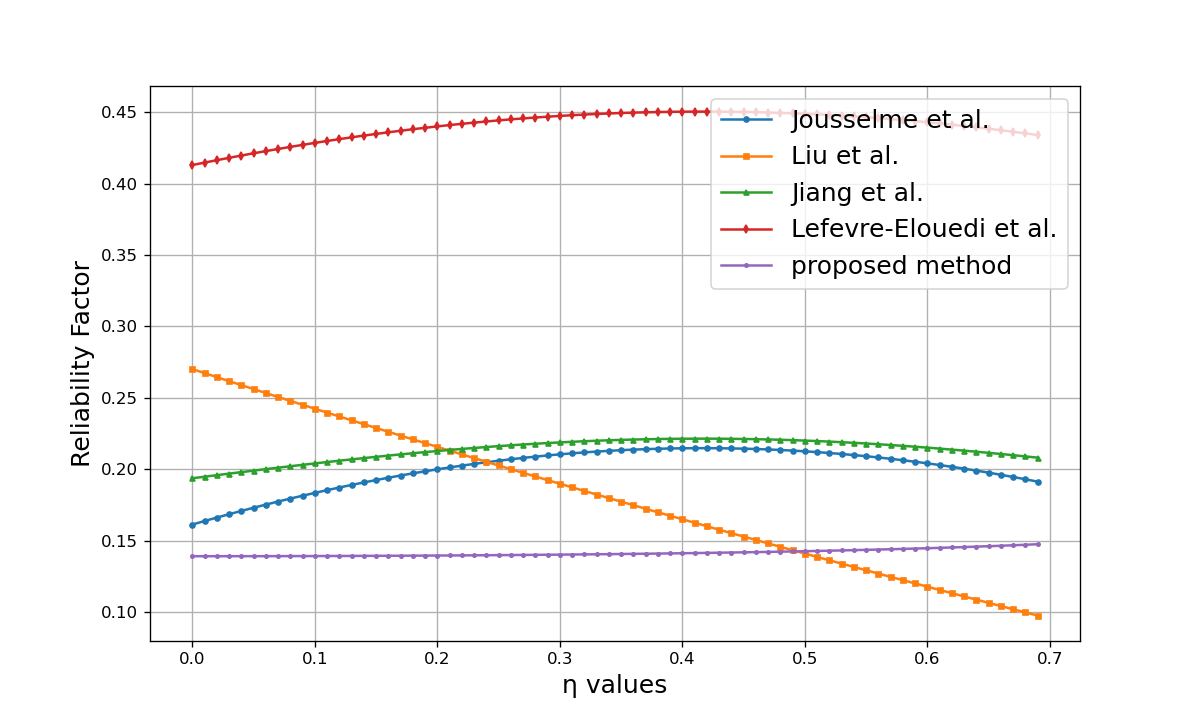} 
    \caption{The reliability factor of information source under different $\eta$.}
    \label{rel2}
\end{figure}

In this example, it can be observed that since $x_1$ is the correct label, the reliability of $m_1$ remains at a relatively low level regardless of how $\eta$ changes. As shown in Figure 4, when $\eta$ continuously changes, the reliability of $m_1$, calculated using the proposed method, does not exhibit significant fluctuations compared to other calculation methods. This is because, in the proposed method, the reliability of $m_1$ is based on both the supporting and opposing Rpt for the correct decision. However, the composition of the opposing evidence is not given much emphasis, making it a calculation method more focused on the final outcome.

\end{example}
\section{Application of the proposed method in target classification}
\label{section 5}
In this section, the method proposed in this paper will be applied to real-world target classification problems. Several comparative algorithms will be used to demonstrate the model's performance and effectiveness in handling classification tasks under varying conditions. These comparisons will highlight the advantages and limitations of the proposed method in relation to existing approaches.

\subsection{Dataset}
The datasets for this experiment are sourced from the UCI Machine Learning Repository, including Iris, Wine, Heart, Australian, Raisin, and Credit Card Clients (CCC). The UC Irvine repository is renowned for its comprehensive collection of datasets, which span various domains and serve as standard benchmarks in the field of machine learning. The specific details of each dataset are presented in \hyperref[dataset_table]{Table 4}.

\begin{table}[ht]
    \centering
    \caption{Summary of Experimental Datasets}
    \resizebox{\linewidth}{!}{
    \begin{tabular}{cccccc}
        \toprule
        & \textbf{Category} & \textbf{Sample Size} & \textbf{Features} & \textbf{Class} & \textbf{Subject Area} \\ \midrule
        1 & Iris & 150 & 4 & 3 & Biology \\
        2 & Wine & 178 & 13 & 3 & Chemistry \\
        3 & Heart & 303 & 13 & 2 & Health and Medicine \\
        4 & Australian & 690 & 14 & 2 & Finance \\
        5 & Raisin & 900 & 7 & 2 & Agriculture \\
        6 & Credit Card Clients & 30,000 & 24 & 2 & Finance \\
        \bottomrule
    \end{tabular}
    }
    \label{dataset_table}
\end{table}

\subsection{Comparative models}
To more comprehensively evaluate the effectiveness of the method proposed in this paper, we selected traditional machine learning models as well as several algorithms based on DS theory.

In the machine learning algorithms selected, we include Decision Tree (DT) \cite{DT}, Support Vector Machine (SVM) \cite{SVM}, Naive Bayes (NaB) \cite{NaB}, K-Nearest Neighbors (KNN) \cite{KNN}, and Logistic Regression (LR) \cite{LR}. Each of these algorithms has distinct characteristics: DT is known for its simplicity and interpretability, SVM is effective in high-dimensional spaces, NaB is efficient for probabilistic classification, KNN is a simple and intuitive instance-based learning method, and LR is widely used for binary classification problems. Together, these algorithms provide a comprehensive comparison of performance, enabling us to evaluate the robustness and generalization of the proposed method across different classification approaches.

In the algorithms based on DS theory, we selected: the traditional DST algorithm \cite{Dempster}, Liu's dissimilarity measurement \cite{Liudissimilarity}, the method proposed by Murphy et al. \cite{Murphyfusion}, the method developed by Deng et al. \cite{Dengfusion}, and the PCA algorithm \cite{PCA}. These methods were chosen for their diverse approaches to handling uncertainty and evidence fusion. The traditional DST algorithm serves as a baseline for comparison, Liu's dissimilarity measurement focuses on evaluating the distance between evidence sources, Murphy's method improves the handling of conflicting evidence, Deng's method introduces a more refined evidence combination strategy, and PCA helps highlight the most significant aspects of the evidence. Together, these DS-based methods provide a balanced perspective for comparing the proposed reliability calculation method in complex decision-making scenarios.

\subsection{Implementation}
\begin{figure*}[h]
    \centering
    \includegraphics[width=0.8\linewidth]{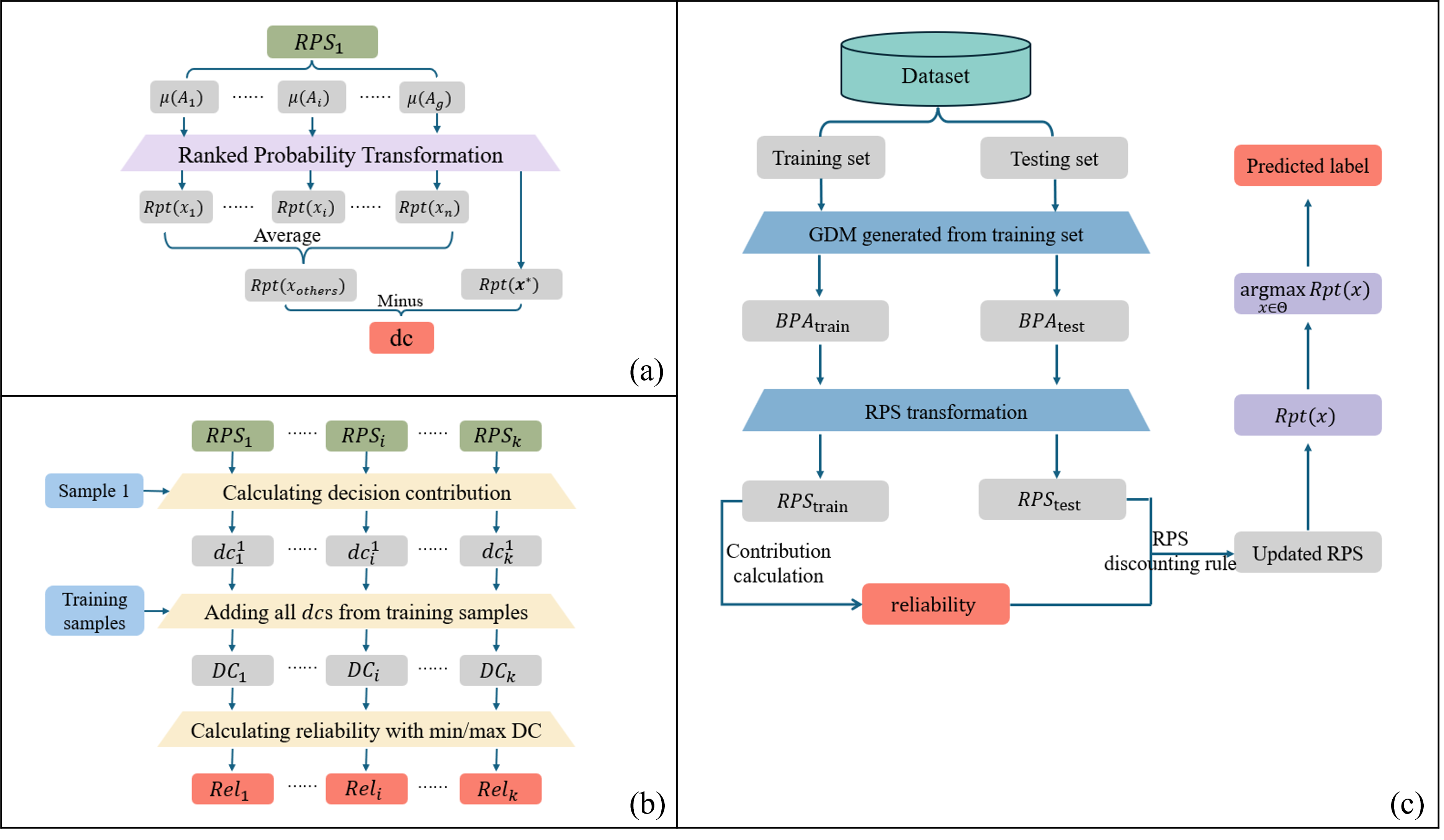} 
    \caption{Overview of the proposed method's diagrams: (a) Calculation of decision contribution, (b) Calculation of evidence reliability, and (c) Classification problem-solving process based on the proposed method.}
    \label{process}
\end{figure*}

\begin{enumerate}[label=\textbf{Step \arabic*:}, leftmargin=1.5cm, labelwidth=1.8cm]
    \item The dataset was randomly divided into training and testing sets using five-fold cross-validation.
    \item In the training set, use the Gaussian discriminant model to generate labeled training BPA, and apply RPS transformation to convert the original BPA into RPS.
    \item Based on the Ranked Probability Transformation, convert different PMFs into probability distributions for different labels.
    \item Calculate the decision contribution of each evidence source toward the correct decision result based on the labels of the training samples, and thereby determine their respective reliability.
    \item In the testing phase, use the Gaussian discriminant model and RPS transformation to obtain the initial BPA and the corresponding RPS for the test samples.
    \item Based on the reliability calculated during the training phase, perform RPS discounting for each evidence source.
    \item Determine the fusion order according to the reliability from highest to lowest, and perform left intersection operations on the RPS sequentially to obtain the final fusion result.
    \item Use the Ranked Probability Transformation to convert the fused PMF into probabilities, obtain the predicted label, and compare it with the correct result.
\end{enumerate}

\subsection{Result and discussion}
\begin{figure}[h]
    \centering
    \includegraphics[width=\linewidth]{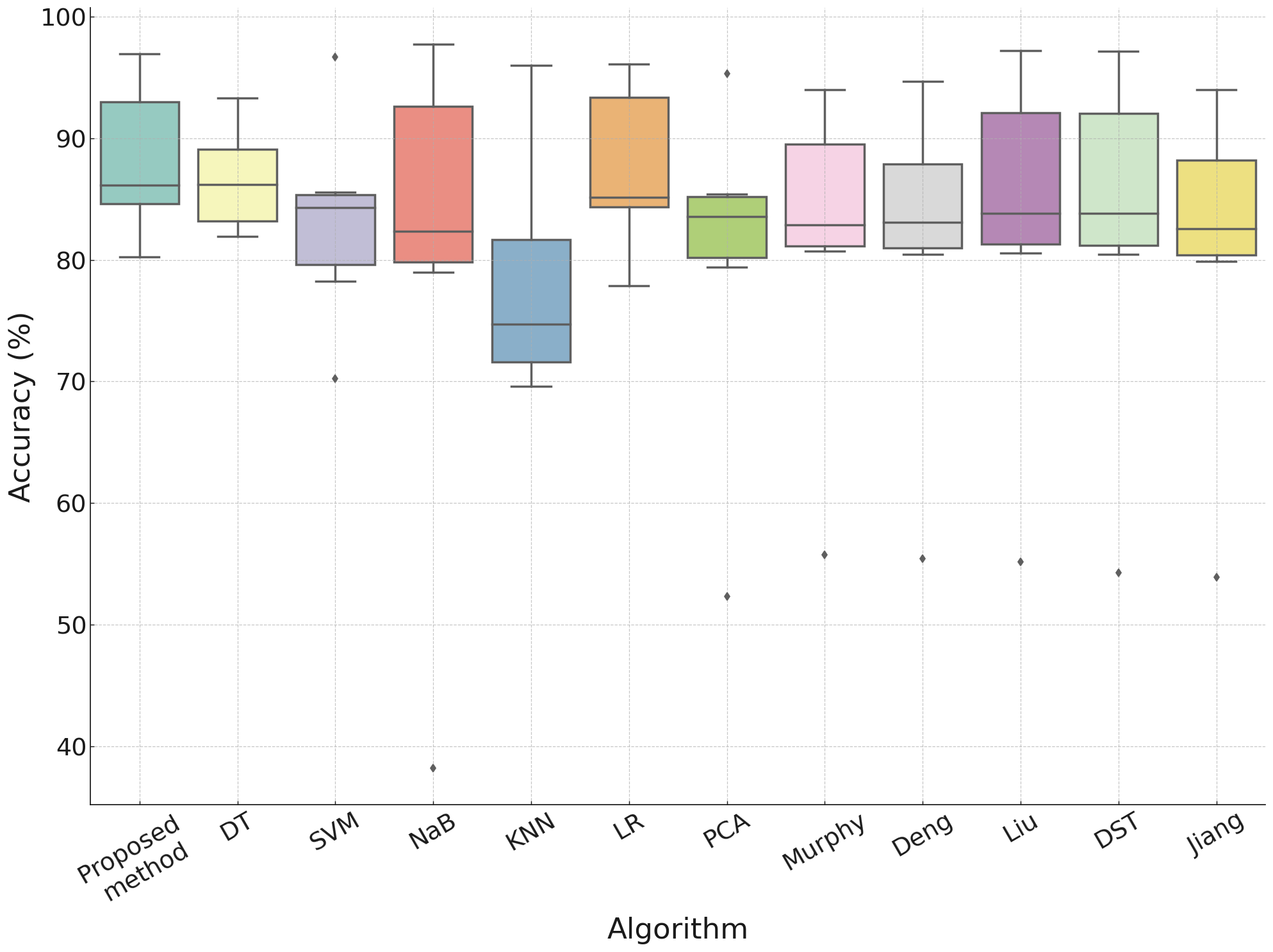} 
    \caption{Classification accuracy of different algorithms on different datasets.}
    \label{result1}
\end{figure}

\begin{figure}[h]
    \centering
    \includegraphics[width=\linewidth]{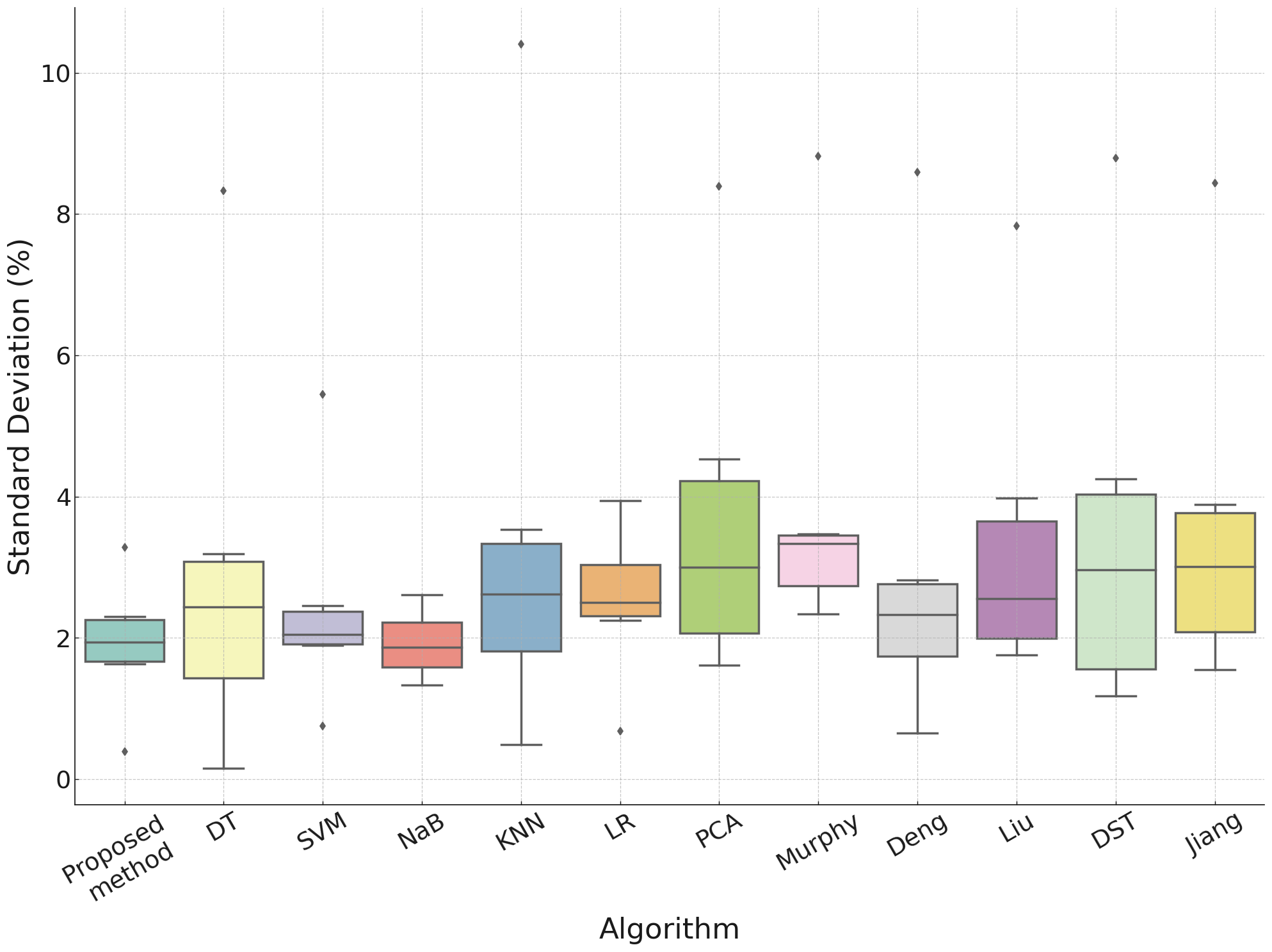} 
    \caption{Classification standard deviation of different algorithms on different datasets.}
    \label{result2}
\end{figure}

In the testing phase, we use five-fold cross-validation to calculate the accuracy and standard deviation of different algorithms. To provide a clearer visualization of the performance of various algorithms across different datasets, we present this data in the form of box plots, as shown in \hyperref[result1]{Fig. 6} and \hyperref[result2]{ 7}. The specific data is detailed in \hyperref[accuracy]{Table 5}.

From \hyperref[result1]{Fig. 6}, it can be seen that the proposed method demonstrates stable overall performance across different datasets, achieving an average accuracy of around 88.14\%, which is higher than most other algorithms. Particularly in the "Iris" and "Wine" datasets, it achieves an accuracy exceeding 95\% with a narrow error range, indicating reliability and consistency in these datasets. Although the accuracy is slightly lower in the "Raisin" and "Australian" datasets, it still maintains good performance and consistency. The proposed method shows notably superior performance in most datasets in comparison with other algorithms, especially in "Iris," "Wine," and "Australian." Moreover, the narrow error range reflects strong robustness across diverse datasets, with stable predictions and superior overall performance.

From \hyperref[result2]{Fig. 7}, the proposed method exhibits stable standard deviation across various datasets, with particularly low variability in the "Iris" and "Wine" datasets, indicating consistent predictive performance. Although the standard deviation is slightly higher in the "Raisin" and "Australian" datasets, it remains within an acceptable range overall. Compared to other algorithms, the proposed method generally maintains lower standard deviation, demonstrating greater predictive consistency and robustness.

In summary, the proposed method demonstrates high efficiency and stability, achieving a relatively high classification accuracy of 88.14\% and a low standard deviation of 1.91\%, outperforming other classification algorithms.

The efficiency and stability of the proposed method are approximately due to the following factors: 1. The reliability calculation is outcome-driven rather than based on the similarity among evidence sources. This approach assigns higher reliability to evidence sources that contribute more to accurate decisions, effectively eliminating interference from highly similar sources that may hinder correct decision-making. 2. The reward and penalty mechanism is designed to ensure that the reliability calculation is not only related to correct decisions but also accounts for decisions that do not support the correct outcome. 3. The RPS transformation meticulously considers the impact of sequence on decision-making, while the Ranked Probability Transformation builds on the RPS transformation by emphasizing the influence of top-ranked predictions on decisions and reducing the interference from lower-ranked labels, thereby enhancing stability. Consequently, the proposed method exhibits strong performance in both accuracy and stability, further demonstrating its superiority.

\begin{table*}[!ht] %
    \centering
    \caption{Classification Accuracy and Standard Deviation Across Datasets}
    \resizebox{\textwidth}{!}{
        \begin{tabular}{lccccccc}
            \toprule
            Algorithm & Iris & Wine & Heart & Australian & Raisin & Credit & Average \\
            \midrule
            DT & 93.33$\pm$2.11 & 89.36$\pm$8.33 & \textbf{88.39}$\pm$1.21 & 82.90$\pm$2.77 & 84.00$\pm$3.19 & \textbf{81.94}$\pm$\textbf{0.15} & 86.65$\pm$2.37 \\
            SVM & 96.67$\pm$ 2.11 & 70.22$\pm$5.45 & 83.80$\pm$1.99 & 84.78$\pm$1.89 & \textbf{85.56}$\pm$2.46 & 78.23$\pm$0.75 & 83.21$\pm$2.44 \\
            NaB & 96.00$\pm$\textbf{1.33} & \textbf{97.76}$\pm$2.08 & 82.24$\pm$2.61 & 78.99$\pm$1.65 & 82.44$\pm$1.56 & 38.17$\pm$2.27 & 79.27$\pm$1.92 \\
            KNN & 96.00$\pm$2.49 & 70.87$\pm$10.41 & 73.85$\pm$3.53 & 69.57$\pm$2.75 & 83.67$\pm$1.59 & 75.59$\pm$0.49 & 78.25$\pm$2.49 \\
            LR & 96.00$\pm$2.49 & 96.08$\pm$2.25 & 84.20$\pm$2.52 & 85.51$\pm$3.94 & 84.78$\pm$3.21 & 77.87$\pm$0.68 & 87.41$\pm$2.52 \\
            \midrule
            DST & 94.67$\pm$3.40 & 97.17$\pm$2.52 & 84.20$\pm$\textbf{1.18} & 80.43$\pm$4.25 & 83.44$\pm$\textbf{1.24} & 54.24$\pm$8.79 & 82.36$\pm$3.23 \\
            PCA & 95.33$\pm$2.67 & 85.38$\pm$4.53 & 82.54$\pm$1.61 & 79.42$\pm$3.32 & 84.56$\pm$1.87 & 52.31$\pm$8.39 & 79.92$\pm$3.06 \\
            Murphy & 94.00$\pm$3.27 & 91.56$\pm$2.56 & 82.34$\pm$2.34 & 80.72$\pm$3.47 & 83.44$\pm$3.40 & 55.74$\pm$8.82 & 81.97$\pm$3.31 \\
            Deng & 94.67$\pm$1.63 & 89.33$\pm$2.05 & 82.63$\pm$2.61 & 80.43$\pm$\textbf{0.65} & 83.56$\pm$2.82 & 55.41$\pm$8.59 & 81.50$\pm$2.73 \\
            Liu & 94.67$\pm$2.67 & 97.19$\pm$\textbf{1.76} & 84.29$\pm$1.84 & 80.58$\pm$2.44 & 83.33$\pm$3.98 & 55.15$\pm$7.83 & 82.53$\pm$2.75 \\
            Jiang & 94.00$\pm$3.89 & 89.87$\pm$3.41 & 82.05$\pm$1.55 & 79.86$\pm$2.61 & 83.11$\pm$1.91 & 53.87$\pm$8.44 & 80.46$\pm$3.02 \\
            Proposed method & \textbf{96.96}$\pm$2.11 & 94.94$\pm$ 3.28 & 85.05$\pm$1.76 & \textbf{87.25}$\pm$1.63 & 84.44$\pm$2.30 & 80.22$\pm$0.39 & \textbf{88.14}$\pm$\textbf{1.91} \\
            \bottomrule
        \end{tabular}
    }
    \label{accuracy}
\end{table*}

\section{Conclusion}
\label{section 6}
This paper introduces a novel approach for evaluating the reliability of evidence sources. The proposed method considers the influence of the internal order of elements within focal sets on decision-making by transforming the BPA into an order-sensitive RPS. Furthermore, based on the varying priorities inherent in the internal order of the RPS, a Ranked Probability Transformation is introduced to emphasize the impact of sequence. During the training phase, the reliability of different RPS sources is calculated based on their defined contribution to correct decision-making.

In the experimental section, several traditional machine learning algorithms and DS-based classification algorithms were selected to compare with the proposed method across multiple datasets. Five-fold cross-validation was used to evaluate and compare the accuracy and standard deviation. The final results demonstrate that the proposed method exhibits strong accuracy and stability in classification tasks.

The main contribution of this paper is the proposal of an RPS transformation method based on BPA, allowing traditional BPA to additionally account for the influence of sequence on decision-making, providing a more refined extension to evidence theory. Meanwhile, the Ranked Probability Transformation is introduced to highlight the significance of sequence, setting it apart from the PPT algorithm. The proposed method is particularly well-suited for supervised learning on large datasets, effectively filtering out unreliable RPS sources that may interfere with decision-making, especially when sample features exhibit high similarity. 

In the future, attention will be paid on integrating the similarity among RPS sources with their contribution to correct outcomes to reduce dependence on the training process. Additionally, establishing more appropriate standards in reliability calculation will be pursued to enhance the applicability of the proposed fusion algorithm.





\bibliographystyle{IEEEtran}
\bibliography{ref}

\newpage

\end{document}